\documentclass{article}

\usepackage{microtype}
\usepackage{graphicx}
\usepackage{subfigure}
\usepackage{booktabs} % for professional tables

\usepackage{hyperref}

\usepackage[accepted]{icml2024}

\usepackage{amsmath}
\usepackage{amssymb}
\usepackage{mathtools}
\usepackage{amsthm}

\usepackage{graphicx}
\usepackage{verbatim}
\usepackage{amsmath}
\usepackage{amssymb}
\usepackage{booktabs}
\usepackage{pifont}
\usepackage{color}
\usepackage{multirow}
\usepackage{overpic}
\usepackage{arydshln}
\usepackage{makecell}
\usepackage{times}
\usepackage{epsfig}
\usepackage[T1]{fontenc}
\usepackage{hhline}
\usepackage{xcolor}
\usepackage{pifont}
\usepackage{enumitem}
\usepackage{colortbl}
\usepackage{verbatim}
\usepackage{bm}
\usepackage{hyperref}
\usepackage{wrapfig}
\def\ie{\textit{i.e.}}
\def\eg{\textit{e.g.}}
\def\etc{\textit{etc}}

\definecolor{darkpastelgreen}{rgb}{0.01, 0.75, 0.24}
\definecolor{darkpink}{rgb}{0.91, 0.33, 0.5}
\definecolor{mygray}{gray}{.92}

\definecolor{linkcolor}{RGB}{255,0,0}
\definecolor{urlcolor}{RGB}{255,105,180}
\definecolor{citecolor}{RGB}{0, 80, 200}
\definecolor{citecolor1}{RGB}{0,153,255}
\usepackage{hyperref}
\hypersetup{colorlinks=true,linkcolor=linkcolor,urlcolor=urlcolor,citecolor=citecolor1}

\usepackage[capitalize]{cleveref}
\crefname{section}{Sec.}{Secs.}
\Crefname{section}{Section}{Sections}
\Crefname{table}{Table}{Tables}
\crefname{table}{Tab.}{Tabs.}

\theoremstyle{plain}

\theoremstyle{definition}

\theoremstyle{remark}

\usepackage[textsize=tiny]{todonotes}

\icmltitlerunning{Submission and Formatting Instructions for ICML 2024}

\begin{document}

\twocolumn[
\icmltitle{Intention-driven Ego-to-Exo Video Generation}



\icmlsetsymbol{equal}{*}
\icmlsetsymbol{corresponding}{$\dagger$}

\begin{icmlauthorlist}
\icmlauthor{Hongchen Luo}{xxx}
\icmlauthor{Kai Zhu}{yyy}
\icmlauthor{Wei Zhai}{xxx,corresponding}
\icmlauthor{Yang Cao}{xxx}

\end{icmlauthorlist}
\icmlaffiliation{xxx}{USTC}
\icmlaffiliation{yyy}{Alibaba Group}
%\icmlaffiliation{yyy}{School of ZZZ, Institute of WWW, Location, Country}

%\icmlcorrespondingauthor{Hongchen Luo}{lhc12@mail.ustc.edu.cn}
\icmlcorrespondingauthor{Wei Zhai}{wzhai056@ustc.edu.cn}
%\icmlcorrespondingauthor{Yang Cao}{forrest@ustc.edu.cn}

%\icmlkeywords{Machine Learning, ICML}

%\icmlaffiliation{yyy}{Department of XXX, University of YYY, Location, Country}
%\icmlaffiliation{comp}{Company Name, Location, Country}
%\icmlaffiliation{sch}{School of ZZZ, Institute of WWW, Location, Country}

%\icmlcorrespondingauthor{Firstname2 Lastname2}{first2.last2@www.uk}

%\icmlkeywords{Machine Learning, ICML}

\vskip 0.3in
]



\printAffiliationsAndNotice{ $^\dagger$Corresponding Author}  % leave blank if no need to mention equal contribution

\begin{abstract}
Ego-to-exo video generation refers to generating the corresponding exocentric video according to the egocentric video, providing valuable applications in AR/VR and embodied AI. Benefiting from advancements in diffusion model techniques, notable progress has been achieved in video generation. However, existing methods build upon the spatiotemporal consistency assumptions between adjacent frames, which cannot be satisfied in the ego-to-exo scenarios due to drastic changes in views.  To this end, this paper proposes an \textbf{I}ntention-\textbf{D}riven \textbf{E}go-to-exo video generation framework (\textbf{IDE}) that leverages action intention consisting of human movement and action description as view-independent representation to guide video generation, preserving the consistency of content and motion. Specifically, the egocentric head trajectory is first estimated through multi-view stereo matching. Then, cross-view feature perception module is introduced to establish correspondences between exo- and ego- views, guiding the trajectory transformation module to infer human full-body movement from the head trajectory. Meanwhile, we present an action description unit that maps the action semantics into the feature space consistent with the exocentric image. Finally, the inferred human movement and high-level action descriptions jointly guide the generation of exocentric motion and interaction content (\ie, corresponding optical flow and occlusion maps) in the backward process of the diffusion model, ultimately warping them into the corresponding exocentric video. We conduct extensive experiments on the relevant dataset with diverse exo-ego video pairs,  and our IDE outperforms state-of-the-art models in both subjective and objective assessments, demonstrating its efficacy in ego-to-exo video generation.
\end{abstract}

\section{Introduction}
\label{introduction}
\begin{figure}[t]
%\vskip 0.2in
	\centering
		\begin{overpic}[width=1.\linewidth]{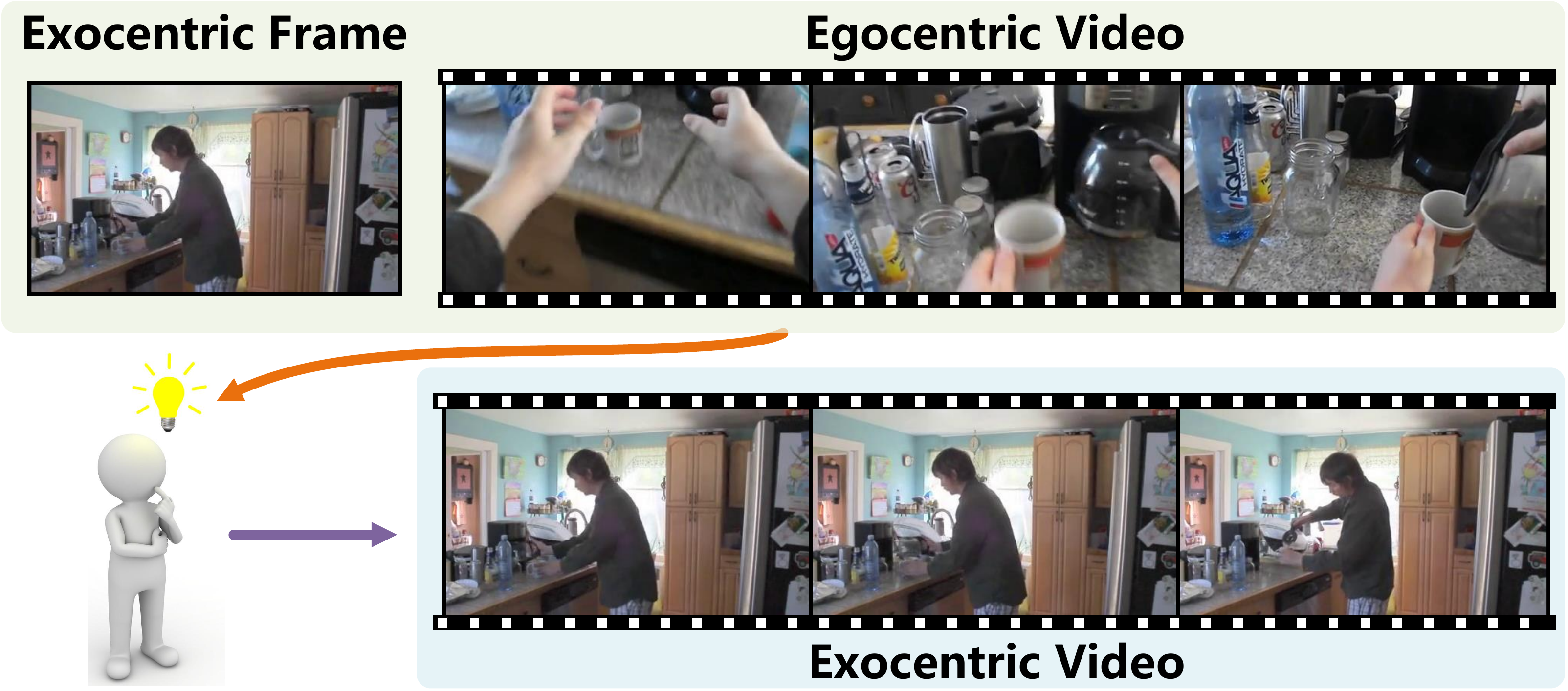}
	\end{overpic}
	\vspace{-11pt}
	\caption{\textbf{Ego-to-exo video generation.} Given an egocentric video and an initial frame of an exocentric, generate an exocentric video for the corresponding scene. }
	\label{fig1}
 \vskip -0.2in
\end{figure}

Generating the corresponding exocentric video using egocentric video (as shown in Fig. \ref{fig1}) enables the model to understand and visualize the same scene from different perspectives\citep{grauman2023ego,wen2021seeing,sigurdsson2018actor}. Due to providing a more comprehensive perception, it has excellent research value in AR/VR,  embodied intelligence, and human-computer interaction \citep{Li_2023_CVPR,rai2021home}.

Benefiting from the development of existing diffusion modeling techniques, video editing and video generation fields have recently achieved significant development \citep{yang2023diffusion}. Most of existing conditional video generation tasks use text/image as a condition to generate the corresponding video \citep{singer2022make,ho2022video,ni2023conditional,ho2022imagen}. Nevertheless, the inherent constraints of textual data pose a challenge in capturing intricate temporal dynamics.  Furthermore, some studies incorporate factors such as human pose, scene semantics, depth maps, \etc., as conditional inputs into the network, providing robust spatiotemporal constraints for video generation \citep{wang2023videocomposer,hu2023videocontrolnet,yin2023dragnuwa}. However, these cues are challenging to acquire between exocentric and egocentric perspectives, making existing methods struggle to address the consistency issues in ego-to-exo video generation. This inconsistency primarily manifests in two aspects: content and motion. Content inconsistency arises from the minimal visual overlap between egocentric and exocentric views. The same object exhibits substantial visual appearance and scale variance across different perspectives. When motion occurs, inconsistency arises from background changes between the egocentric view and variations in the activity scene content observed from the exocentric perspective.
%arise between background changes in the egocentric view and variations in the activity scene content observed from the exocentric perspective. 
Motion inconsistency is evident because of the simultaneous movements of a human's body and the egocentric camera. Discrepancies exist in the direction of head movement, relative velocities concerning body motion, and changes in pose. These challenges make it difficult to generate exocentric videos that correspond appropriately to egocentric videos.

\begin{figure}[t]
	\centering
		\begin{overpic}[width=0.94\linewidth]{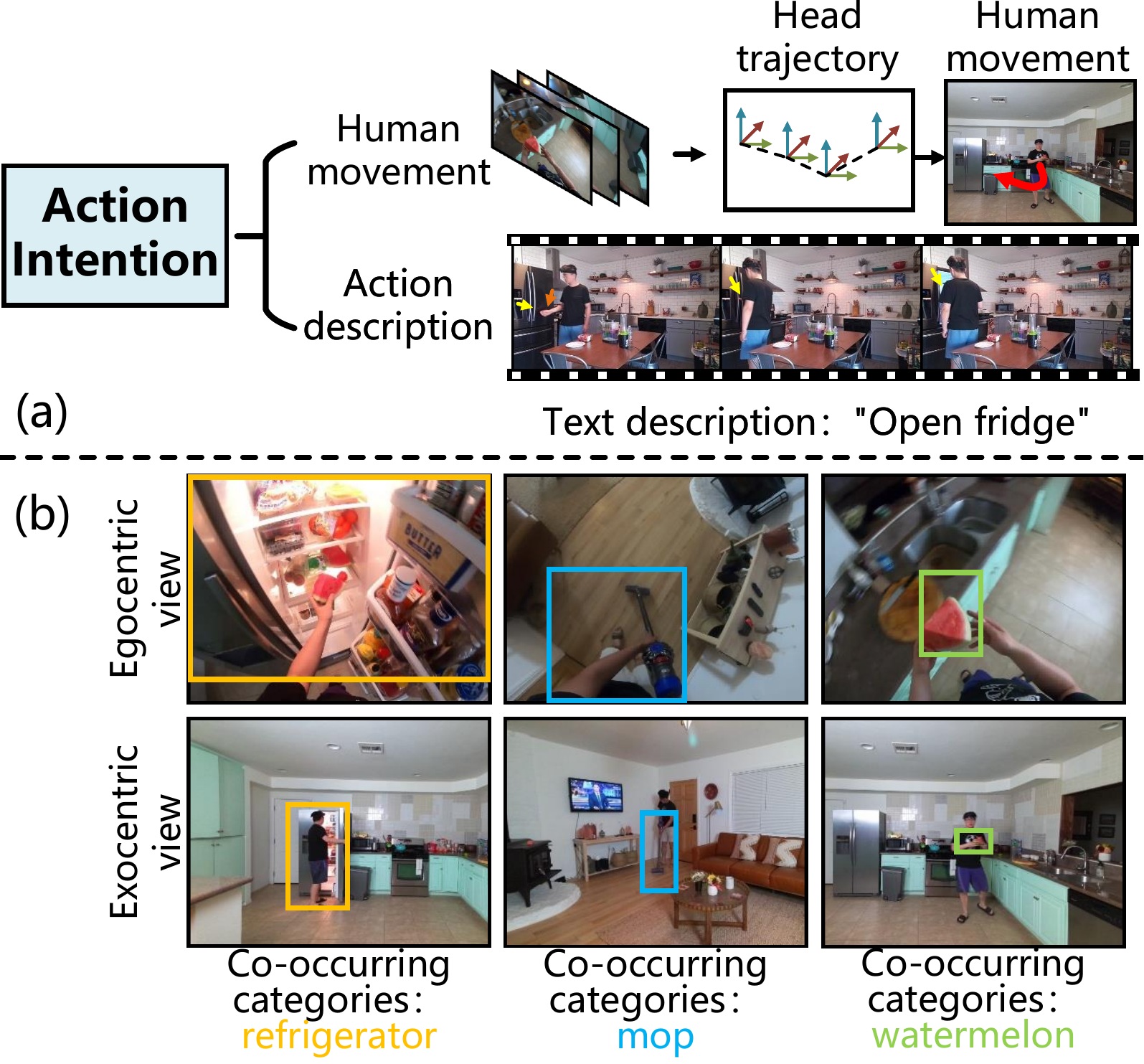}
	\end{overpic}
	\vspace{-8pt}
	\caption{\textbf{Motivation.} \textbf{(a)} The human action intention consists of the human movement and the action description. The human movement can be obtained indirectly from the head trajectory. \textbf{(b)} Utilizing objects co-occurring in the exocentric and egocentric, it is feasible to establish a connection between the two perspectives and achieve content alignment.}
	\label{motivation}
 \vskip -0.2in
\end{figure}

Between the two perspectives,  human intention acts as a view-independent invariant representation, serving as a bridge to transfer motion from egocentric to exocentric views and mitigating content and motion inconsistencies arising from varying perspectives. Motivated by it, we consider utilizing action intention as an intermediary representation between the two perspectives, assisting in generating exocentric videos consistent with egocentric ones. Human action intention can be described by both human movement and 
action description. As shown in Fig. \ref{motivation} (a), the human movement provides crucial clues about the relative positional changes of the human in the scene (\ie, rotation and translation), while action descriptions offer high-level information guiding interactions within the scene (\eg, completing the action of opening by touching a handle with a hand). The human movement is challenging to obtain directly. Considering this, we explore leveraging the potential connection between head and human motions, where the head motion trajectory provides the approximate direction and translation of human motion, enabling the indirect inference of an approximate human movement. Additionally, we consider using class tokens as a bridge connecting the two views (as shown in Fig. \ref{motivation} (b)). This guidance helps the model explore objects in both egocentric and exocentric views, establishing content alignment between the two viewpoints and facilitating the conversion of the head motion trajectories into approximate human movement.

In this paper, we propose a novel \textbf{I}ntention-\textbf{D}riven \textbf{E}go-to-exo video generation framework (\textbf{IDE}) (as shown in Fig. \ref{pipeline}). Firstly, a cross-view feature perception module (\textbf{CFPM}) is introduced to utilize class tokens to query features from the other viewpoint to establish cross-view content correspondences. The class token is then used to adjust the feature distribution within the local viewpoint, directing the model to focus on regions where content coherence exists in both perspectives. Subsequently, a trajectory transformation module (\textbf{TTM}) is devised to utilize head motion trajectories to adjust the temporal feature distribution of the egocentric viewpoint, imbuing it with temporal motion cues. Leveraging the connections established by the CFPM, these cues are transferred to the exocentric viewpoint, allowing dynamic adjustments in feature distribution based on motion cues. Simultaneously, an action description unit (\textbf{ADU}) is presented to map the action semantics into the feature space consistent with the exocentric image. Finally, the adjusted exocentric feature distribution and text feature serve as conditions input to the backward process of the diffusion model, guiding the generation of the corresponding optical flow and occlusion maps, which finally obtain the exocentric video through the warp transformation.

The main contributions of this paper are summarised as follows: \textbf{1)} We propose an \textbf{I}ntention-\textbf{D}riven \textbf{E}go-to-exo video generation framework (\textbf{IDE}). IDE initially converts egocentric head trajectory into human movement. By jointly utilizing human movement and action description to represent human action intention, IDE assists the network in generating motion videos within the exocentric scenario. \textbf{2)} We introduce a novel cross-view feature perception module (CFPM) that provides inter-view content alignment for head trajectories converted to human movement by mining objects semantically shared between exocentric and egocentric views.
\textbf{3)} We conduct experiments on the relevant dataset containing rich exo-ego video pairs, and the experimental results proved that our proposed method outperforms state-of-the-art video generation models.

\section{Related Work}
\subsection{Cross-view Learning}
Exocentric view video captures human action pose and environmental contextual information, while egocentric view focuses on human-object interactions and the operator's attention \citep{grauman2023ego,jia2020lemma}. Both are equally important for capturing human action skills and understanding human-object interactions in the environment. Recently, a large number of researchers focused on tasks such as cross-view feature alignment \citep{li2021ego,xue2023learning,Wang_2023_ICCV,truong2023cross}, video summarization \citep{ho2018summarizing}, image/video retrieval \citep{fan2017identifying}. 
Another study considers cross-view generation, which is divided into generating a human pose based on an egocentric \citep{akada20233d,Li_2023_CVPR} and generating a corresponding egocentric video based on an exocentric video \citep{liu2021cross}. In contrast to the above work, we convert egocentric head trajectories to human movement and later use the human movement and action descriptions together to represent the action intention to assist the network in generating exocentric videos consistent with the egocentric videos.

\subsection{Conditional Video Generation}
Early research in video generation \citep{hong2022cogvideo,skorokhodov2022stylegan} focused on using GAN-related methods, but maintaining spatio-temporal coherence and realism remains a major challenge. Since the emergence of the diffusion model, several SOTAs have emerged in generative modeling and significant progress has been made in several areas such as image/video generation \citep{yang2023diffusion,rombach2022high,Ruiz_2023_CVPR}.
Conditional video generation refers to generating corresponding videos based on user-provided signals. Existing work focuses on text-to-video \citep{singer2022make,Wu_2023_ICCV}, image-to-video \citep{ni2023conditional,wang2023videocomposer}, and motion-guided conditional video generation \citep{hu2023videocontrolnet,ma2023trailblazer}. However, these works build upon the spatiotemporal consistency assumptions between adjacent frames. In contrast, in ego-to-exo video generation, the above condition cannot be satisfied due to the drastic changes in viewpoints. To this end, we utilize action intentions as viewpoint-independent invariant representations to maintain content and motion consistency between exocentric and egocentric viewpoints.

\section{Method}
\begin{figure}[t]
	\centering
		\begin{overpic}[width=1.\linewidth]{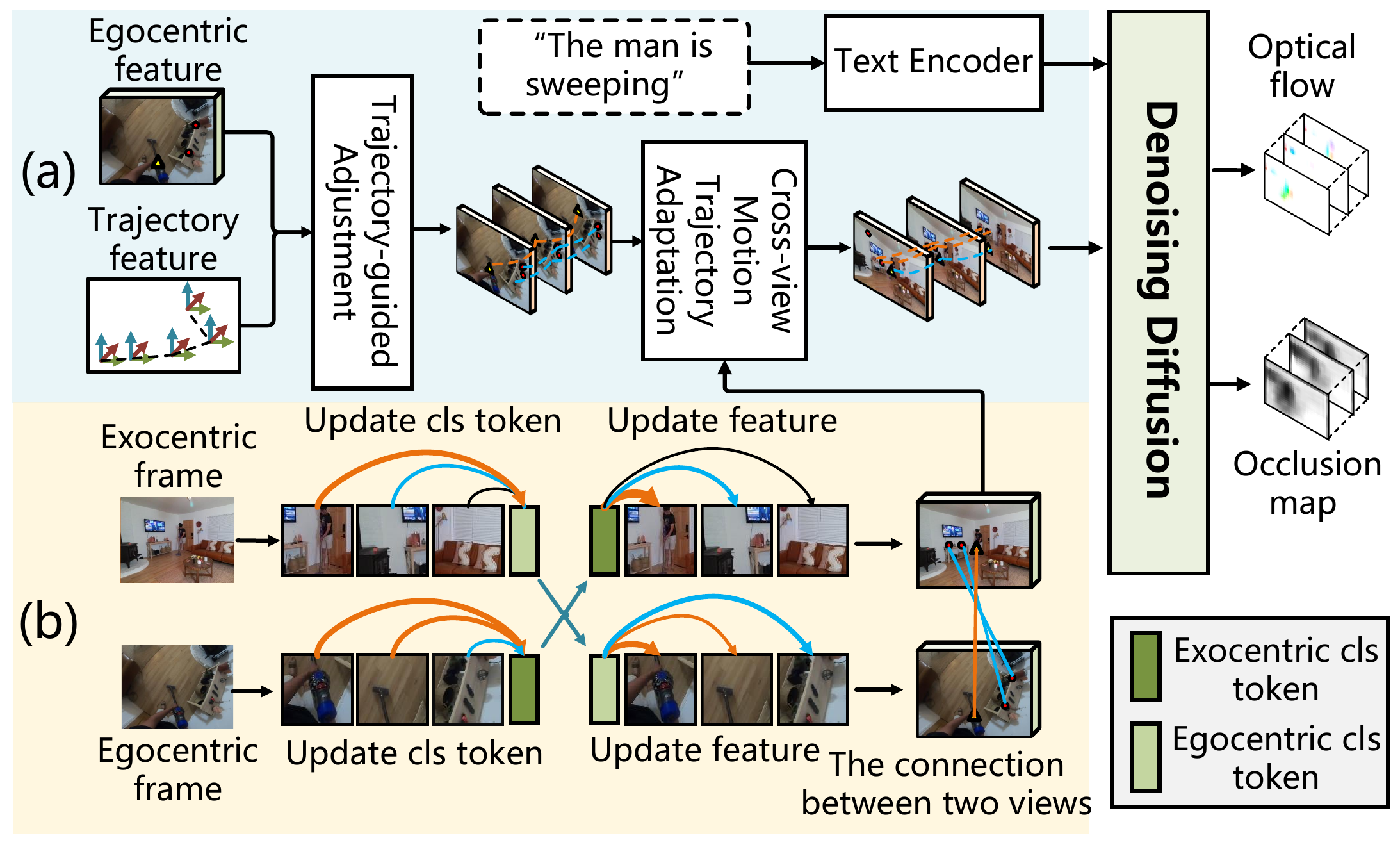}
	\end{overpic}
	\vspace{-2pt}
	\caption{\textbf{The ego-to-exo video generation pipeline.} (a) Human movement is inferred from the head trajectory and the relationship between the two views, while the text encoder maps the action description to the feature space consistent with the exocentric image. These two components serve as conditional inputs to the backward process of the diffusion model, guiding the generation of corresponding optical flow and occlusion maps. (b) Mining of objects shared by different viewpoints to establish content alignment between the two views. }
	\label{pipeline}
 \vskip -0.2in
\end{figure}

Given an egocentric video $\mathcal{V}^{ego}=\{I^{ego}_1,...,I_T^{ego}\}$ ($T$ is the number of video frames) and an exocentric video first frame $I^{exo}_1$, the goal is to generate the corresponding exocentric video sequence. Our pipeline is shown in Fig. \ref{pipeline}. Firstly, the cross-view feature perception module (CFPM) is introduced to establish the connection between egocentric and exocentric video frames. Subsequently, the trajectory transformation module (TTM) utilizes the connection between ego-to-exo to transfer the head motion information to the exocentric features. Meanwhile, the action description unit (ADU) is use to map the action semantics into the feature space consistent with the exocentric image, which is together with the adjusted exocentric feature fed into the diffusion model as the conditions for generating the corresponding optical flow and occlusion maps. Finally, using the warp transform to obtain the exocentric video outputs.

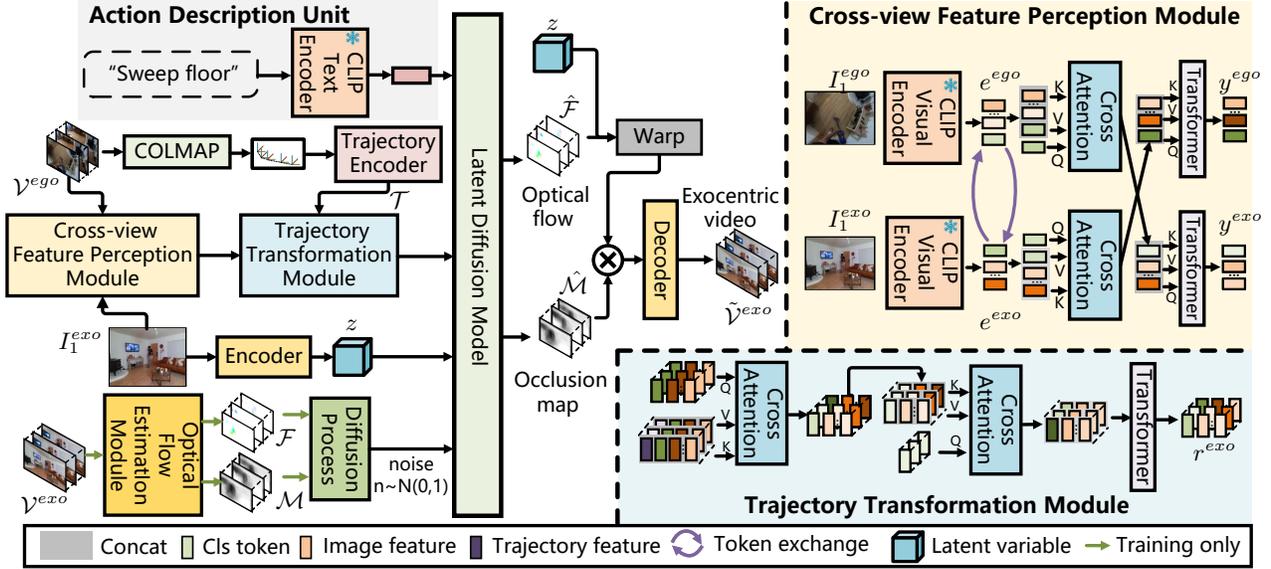
\begin{figure*}[t]
	\centering
		\begin{overpic}[width=0.98\linewidth]{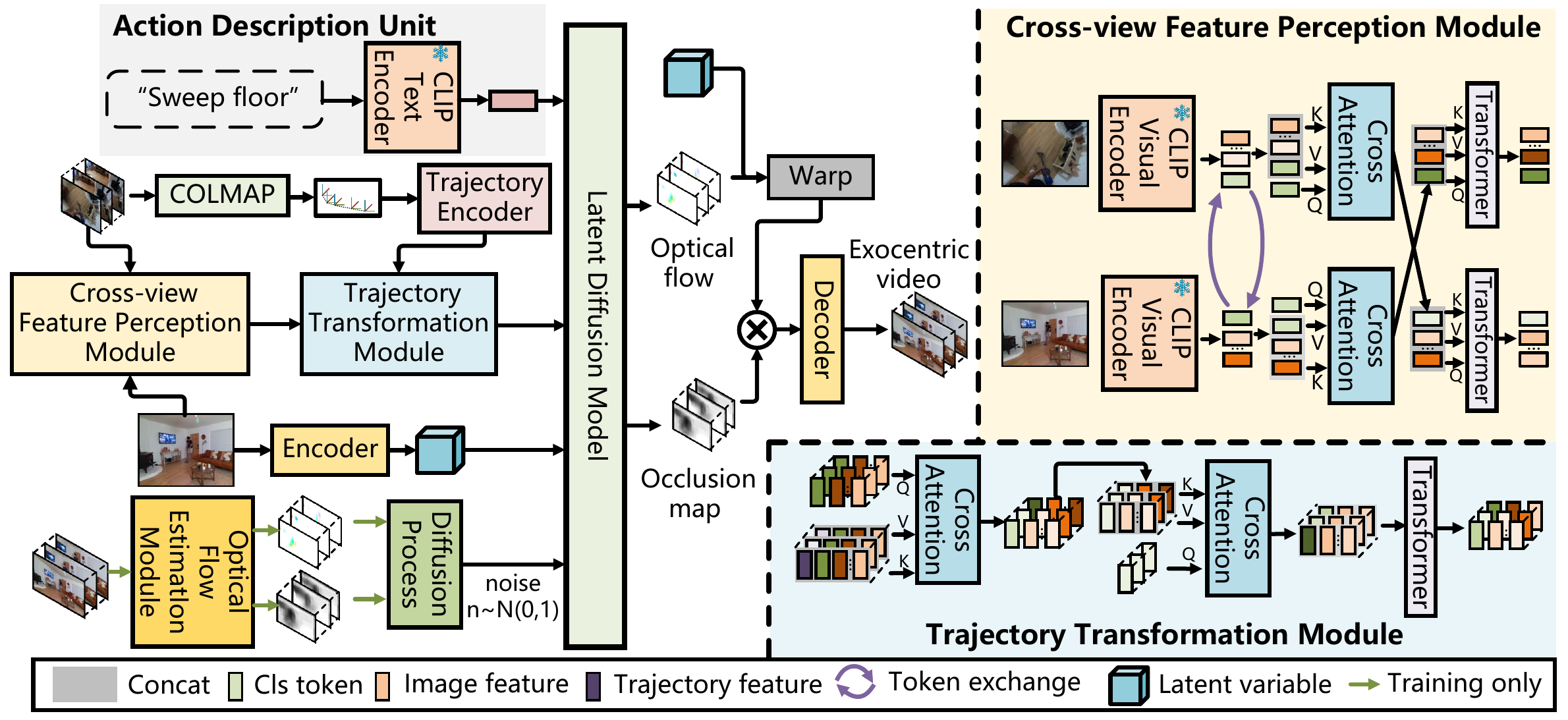}
  \put(1.2,30.){\small{$\mathcal{V}^{ego}$}}
  %\put(20.3,19.9){\small{$\phi$}}
  %\put(51.3,18.1){\small{$\Omega$}}
	 \put(27.5,19.3){\small{$z$}} 
   \put(22.,10.2){\small{$\mathcal{F}$}}
   \put(22.,5.2){\small{$\mathcal{M}$}}
    \put(44.4,36.2){\small{$\hat{\mathcal{F}}$}}
   \put(44.4,22.2){\small{$\hat{\mathcal{M}}$}}
   \put(31,29.){\small{$\mathcal{T}$}}
   \put(96.6,38.4){\small{$y^{ego}$}}
   \put(96.6,27.2){\small{$y^{exo}$}}
   \put(94.6,9.){\small{$r^{exo}$}}
   \put(57.5,19.8){\small{$\tilde{\mathcal{V}}^{exo}$}}
   %\put(32.28,26.){\small{$r^{exo}$}}
   \put(77.7,19.4){\small{$e^{exo}$}}
   \put(77.7,38.4){\small{$e^{ego}$}}
   \put(65.7,27.5){\small{$I^{exo}_1$}}
   \put(65.7,38.5){\small{$I^{ego}_1$}}
   \put(1.8,4.8){\small{$\mathcal{V}^{exo}$}}
   \put(4.8,17.8){\small{$I^{exo}_1$}}
  \put(43.4,43.1){\small{$z$}} 
	\end{overpic}
	\vspace{-5pt}
	\caption{\textbf{\textbf{I}ntention-\textbf{D}riven \textbf{E}go-to-exo video generation framework (\textbf{IDE}).} The cross-view feature perception module (CFPM) uses class tokens from different viewpoints to mine objects common to exocentric and egocentric video frames to establish connections between regions between ego-to-exo view. The trajectory transformation module first utilizes the head motion to adjust the dynamic distribution of the egocentric features temporally. Then, it leverages the ego-to-exo connections established by the CFPM to transfer the motion information to the exocentric features and the action description to provide more accurate interaction cues.}
	\label{framework}
 \vskip -0.2in
\end{figure*}

The framework is shown in Fig. \ref{framework}, with the whole process divided into two stages: the first stage is training a latent flow  auto-encoder \citep{ni2023conditional}, and the second stage is training a conditional diffusion model. 
The network in the first stage consists of three main components: an image encoder for mapping the image to the latent variable, an optical flow estimation module for estimating the latent flow $f$ and occlusion $m$, and a decoder for warped latent maps to the final output $\tilde{z}$. During the first phase of the training process, the network randomly selects two frames ($I_i$ and $I_j$, $i\neq j$) from the same video, then inputs $I_i$ into the encoder to acquire the latent variable $z$. Meanwhile, $I_i$ and $I_j$ are input into the optical flow estimation module to obtain the backward optical flow $f$ and occlusion $m$ from $I_j$ to $I_i$. From $z$, $f$ and $m$, we can obtain the latent variable $\tilde{z}$ for warped: $\tilde{z}=m\otimes\mathcal{W}(z,f)$, where $\mathcal{W}()$ denotes backward warped map and $\otimes$ represents element-wise multiplication. Finally, the decoder decodes $\tilde{z}$ to the final output $\tilde{I}_2$. The training of the model relies on the reconstruction of the loss: $\mathcal{L}_{rec}=||\tilde{I}_j-I_j||^2$. Furthermore, we added the perceptual loss ($\mathcal{L}_{per}$) based on the VGG \citep{simonyan2014very}, and the total loss is denoted as: $\mathcal{L}_{stage1}=\mathcal{L}_{rec}+\lambda\mathcal{L}_{per}$. $\lambda$ represents the hyper-parameters of the perceptual loss weights.
The second stage freezes the parameters of the latent optical flow auto-encoder trained in the first stage to train a conditional video generation model, which mainly contains a cross-view feature perception module (CFPM) (see in Sec. \ref{CFE}), an action description unit (ADU), a trajectory transformation module (TTM) (see in Sec. \ref{TTM}), and a latent diffusion model (see in Sec. \ref{dm}). Specifically, the first frames of the exocentric and egocentric ($I_1^{exo}$ and $I_1^{ego}$) are firstly fed into the CFPM together to establish the connection between the exocentric and egocentric views. Then, the output and the head motion trajectory are fed into the TTM  to dynamically adjust the exocentric feature representation $r^{exo}$. Meanwhile, the ADU extracts the action description features $t$ using the CLIP \citep{radford2021learning} text encoder. Finally, the adjusted exocentric feature representation $r^{exo}$, the latent variable $z$, and the action semantic features $t$ are jointly fed into the latent diffusion model as conditions to predict the latent variables of optical flow and occlusion map.

\subsection{Cross-view Feature Perception Module}
\label{CFE}
The cross-view feature perception module (CFPM) primarily employs class tokens to guide the network in extracting objects that co-occur in both viewpoints, facilitating the establishment of the content connection between exocentric and egocentric views. For the input $I^{exo}$ and $I^{ego}$, the features are first extracted using the CLIP visual encoder \citep{radford2021learning} to obtain $e^{exo}$ and $e^{ego}$, respectively. Subsequently, we utilize cross-attention to enhance the cross-view feature representation of the class token. Take the egocentric feature branch as an example, we first utilize the egocentric feature's class token $f_{cls}^{ego}$ as a query along with the exocentric feature to input into the cross attention for cross-view information fusion. The exocentric feature branch performs a similar operation. The operation of the cross-attention $y=CA(e_1,e_2)$ is as follows:
\begin{equation}
    \small
    d=[e_1,e_2], 
\end{equation}
\begin{equation}
    \small
    q=e_1W_q,\quad k=dW_k,\quad v=dW_v,
\end{equation}
\begin{equation}
    \small
    A=softmax(qk^T/\sqrt{C/h}, \quad Y=Av+e_1.
\end{equation}
where $C$ and $h$ are the feature dimension and head number, $[\cdot , \cdot]$ means the concatenation operation. For the input $e_{exo}$, $e_{ego}$, the cross-view information fusion operates as follows:
\begin{equation}
    \small
y_{cls}^{ego}=CA(e^{ego}_{cls},e^{exo}_{patch}), \quad y_{cls}^{exo}=CA(e^{exo}_{cls},e^{ego}_{patch})
\end{equation}
where $f_{cls}$ and $f_{patch}$ represent the class token and  patch features, respectively. To ensure that cross attention can exploit object regions common to both views, we introduce an alignment loss that ensures that the class tokens of the exocentric view are distributed as closely matched to the egocentric as possible: $\mathcal{L}_{align}=KL(y^{exo}_{cls}||y^{ego}_{cls})$.
After updating the class token from cross views, we use the transformer layer \citep{dosovitskiy2020image} to update the patch features in the respective viewpoints:
\begin{equation}
\small
   y^{ego}=Transformer([y_{cls}^{ego},e^{ego}_{patch}]).
\end{equation}
\begin{equation}
\small
   y^{exo}=Transformer([y_{cls}^{exo},e^{exo}_{patch}]).
\end{equation}

\subsection{Trajectory Transformation Module}
\label{TTM}
The trajectory transformation module mainly considers the dynamic connection between the egocentric and exocentric views in the temporal sequence using the head's motion trajectory adjustment. As shown in Fig. \ref{framework}, we first acquire the head trajectory of the egocentric using colmap \citep{schoenberger2016sfm} and extract the trajectory's features using a trajectory encoder that project it to be aligned with the $y^{ego}$ dimension. Subsequently, the egocentric features are replicated $T$ times along the time dimension to obtain a static feature sequence (\ie, $R^{T \times C \times L}, L=H \times W$), and the trajectory features are jointly fed into cross-attention with the egocentric features to perform feature fusion on them in the temporal dimension:
\begin{equation}
    \small
    r^{ego}=CA(y^{ego},\mathcal{T}),
\end{equation}
Then, the class token in exocentric is utilized as a query to go through a cross-attention layer with the exocentric features for dynamic information fusion across perspectives:
\begin{equation}
   \small r^{exo}_{cls}=CA(y^{exo}_{cls},r^{ego}_{patch}).
\end{equation}
Finally, we concatenate the $r^{exo}_{cls}$ and $y^{exo}_{patch}$ and send them to a transformer layer to complete the dynamic updating of the exocentric features ($r^{exo}$):
\begin{equation}
    \small  r^{exo}=Transformer([r^{exo}_{cls},y^{exo}_{patch}]).
\end{equation}

\subsection{Exocentric Video Generation}
\label{dm}
Since the dynamic connection between exocentric and egocentric and the like is relatively coarse, and the trajectories obtained by colmap are not completely accurate, in order to obtain accurate exocentric video sequences, it is still necessary to utilize the scene context clues in exocentric, both to the existing conditions to generate a model inspired by the model, we deployed a latent optical flow diffusion model \citep{ni2023conditional} to generate the optical flow under the exocentric viewpoint. We use DDPM to construct probabilistic diffusion models \citep{ho2020denoising}. The main idea of the diffusion model is to design forward processes that add Gaussian noise of known variance to the original data and learn a denoising model enabling progressive denoising to generate $x_0$ from a normal distribution sampled $x_N$. Specifically, the forward diffusion process can be expressed using an $N$ step Markov chain as:
\begin{equation}
    \small
    q(x_{1:N}|x_0):=\prod_{n=1}^N q(x_n|x_{n-1}).
\end{equation}
Each step is decided by the variance schedule $\beta_n$:
\begin{equation}
    \small
    q(x_n|x_{n-1}):=\mathcal{N}(x_n;\sqrt{1-\beta_n}x_{n-1},\beta_n\textbf{I}).
\end{equation}
In the inverse process, we consider the conditional generation approach, whose process can be approximated as a Markov chain with a learned mean and fixed variance:
\begin{equation}
    \small
    p_{\epsilon}(x_{n-1}|x_n,c):=\mathcal{N}(x_{n-1}:\mu_\epsilon(x_n,n,c),\sigma^2_n\textbf{I}),
\end{equation}
where $\epsilon$ is the parameters of denoising model, In this module, $\epsilon$ is implemented by time-conditional U-Net \citep{ronneberger2015u} with residual blocks \citep{he2016deep} and self-attention layers \citep{vaswani2017attention}. Where $x_0$ denotes the input after optical flow and occlusion map concatenation. $c$  jointly represented by the TTM output $r^{exo}$, action description feature $t$ and the hidden variable $z$. The learned mean $\mu_\theta$ can be represented as follows:
\begin{equation}
    \small
    \mu_\epsilon=\frac{\sqrt{\alpha}_n(1-\bar{\alpha}_{n-1}+\sqrt{\bar{\alpha}_{n-1}}(1-\alpha_n)\hat{x}_\epsilon(x_n,n,c)}{1-\bar{\alpha}_n}.
\end{equation}
The loss of the training process can be defined as:
\begin{equation}
    \small
\mathcal{L}_{dm}=E_{x_0,n}||\hat{x}_\epsilon(x_n,n,c)-x_0||.
\end{equation}
Finally, the second stage of the loss consists of the $\mathcal{L}_{align}$ and $\mathcal{L}_{dm}$: $\mathcal{L}_{stage2}=\mathcal{L}_{align}+\mathcal{L}_{dm}$.

\begin{table*}[t]
\caption{\textbf{The results of different methods on ego-to-exo video generation.} The experiments compare the ImaGINator \citep{wang2020imaginator}, VDM \citep{ho2022video} and LFDM \citep{ni2023conditional} methods. The subscript $64$ represents the test results of LFDM and our method at a resolution of $64 \times 64$. \textbf{Bold} represents the best results under each metric.}
\label{Table:total result}
%\vskip 0.05in
\begin{center}
\renewcommand{\arraystretch}{1.}
  \renewcommand{\tabcolsep}{14pt}
\begin{small}
%\begin{sc}
 \begin{tabular}{r||ccc|ccc}
%\toprule
\hline
    \Xhline{2.\arrayrulewidth}
\multirow{2}{*}{\textbf{Method}}    & \multicolumn{3}{c|}{\textbf{\texttt{Seen}}} & \multicolumn{3}{c}{\textbf{\texttt{Unseen}}} \\
  \cline{2-7}
   &    $\text{LPIPS} \downarrow$  & $\text{FVD} \downarrow$ & $\text{KVD} \downarrow$  &  $\text{LPIPS} \downarrow$   & $\text{FVD} \downarrow$ & $\text{KVD} \downarrow$ \\ 
 \hline
    \Xhline{2.\arrayrulewidth}
  ImaGINator \citep{wang2020imaginator} &  $0.546$ & $544.23$ & $34.60$  & $0.546$ & $657.21$ & $39.74$ \\
  %ImaGINator \\
  VDM \citep{ho2022video}  &  $0.342$ & $323.54$  & $16.25$  & $0.338$ & $396.16$ & $19.36$\\
  %VDM \\ 
  LFDM$_{64}$ \citep{ni2023conditional}  &  $0.113$& $308.63$  & $22.39$ & $0.092$ & $403.53$ & $21.84$ \\
  LFDM \citep{ni2023conditional}  & $0.191$ &  $342.82$ & $42.42$  & $0.184$  & $420.74$ & $30.99$ \\
 
  \hline
  \rowcolor{mygray}
  Ours$_{64}$  & $\textbf{0.054}$ & $\textbf{190.52}$ & $\textbf{12.35}$  & $\textbf{0.084}$ & $\textbf{248.98}$ & $\textbf{13.59}$ \\
  \rowcolor{mygray}
  Ours & $\textbf{0.133}$ & $\textbf{219.81}$ & $\textbf{22.49}$  & $\textbf{0.177}$ & $\textbf{310.96}$ & $\textbf{24.13}$ \\
\hline
    \Xhline{2.\arrayrulewidth}
\end{tabular}
%\end{sc}
\end{small}
\end{center}
\vskip -0.1in
\end{table*}

\begin{figure*}[t]
	\centering
		\begin{overpic}[width=0.985\linewidth]{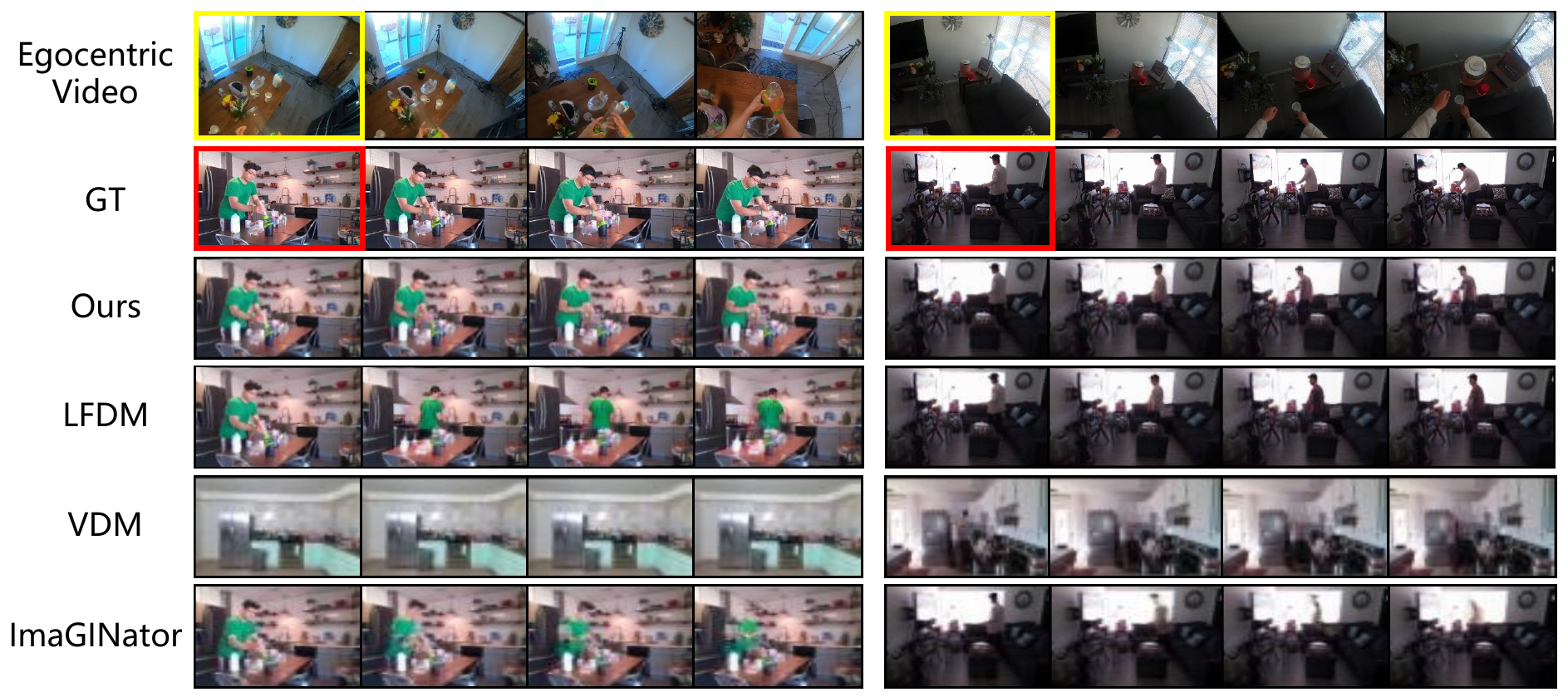}
	\end{overpic}
	\vspace{-10pt}
	\caption{\textbf{The results for exocentric video generation with different methods in \texttt{Seen} setting.} The \textcolor{yellow}{yellow} box represents the first frame of the egocentric video and the \textcolor{red}{red} box represents the first frame of the real exocentric video.}
	\label{result}
 \vskip -0.2in
\end{figure*}

\subsection{Inference}
The inference phase is shown in Fig. \ref{framework}. Given an egocentric video $V^{ego}$, the action description and the first frame $I_1^{exo}$ of the exocentric video, the image encoder $\phi$ encodes $I_1^{exo}$ as $z$. At the same time, $I_1^{exo}$ and $I_1^{ego}$ are fed into the CFPM to obtain the cross-view enhanced features $y^{exo}$ and $y^{ego}$. Subsequently, they are fed into the trajectory transformation module to obtain the dynamically adjusted features $r^{exo}$. And the action description is fed to the ADU to obtain the feature representation $t$. Finally, $r^{exo}$, action description features $t$, $z$, and randomly sampled noise $n$ from Gaussian noise are fed together into the network $\epsilon$, which is progressively denoised by the DDPM to produce the latent flow sequences and $\mathcal{M}=\{m_1,...,m_T\}$. Subsequently, $\mathcal{F}$ and $\mathcal{M}$ are warped transformed and then fed into the decoder to obtain the final exocentric video $\mathcal{\tilde{V}}^{exo}$.

\section{Experiment}
\subsection{Experimental Setup}
\textbf{Datasets. \ } We choose LEMMA \citep{jia2020lemma}, an exo-ego view-aligned video dataset containing diverse and complex scenarios and a wide range of human activities. Our focus is on videos depicting an agent engaged in tasks and establish two divisions: (1) \textbf{\texttt{Seen}}: we cut the long videos according to the action annotation and directly divides all video clips into training and testing sets according to 8:2, (2) \textbf{\texttt{Unseen}}: we divides all long videos into training and testing sets according to 8:2, then cut each long video into short clips according to the annotation. The first division mainly evaluates the model's ability to predict different motions for the same scene, while the second division evaluates the model's generalization performance. 

\textbf{Metrics.\ } To comprehensively evaluate the performance of the different models, we choose three widely used metrics: Learned Perceptual Image Patch Similarity (LPIPS) \citep{zhang2018unreasonable},  Fréchet Video Distance (FVD) \citep{unterthiner2018towards} and Kernel Video Distance (KVD) \citep{unterthiner2018towards}. LPIPS serves as a metric for assessing the perceived similarity between two images, while FVD and KVD evaluate the quality of generated videos. FVD extracts feature from video clips using pre-trained I3D \citep{carreira2017quo}, computing scores from mean and covariance matrices. Similar to FVD, KVD utilizes I3D features and evaluates video quality through the kernel-based Maximum Mean Difference (MMD) method.

\begin{figure*}[t]
	\centering
		\begin{overpic}[width=0.985\linewidth]{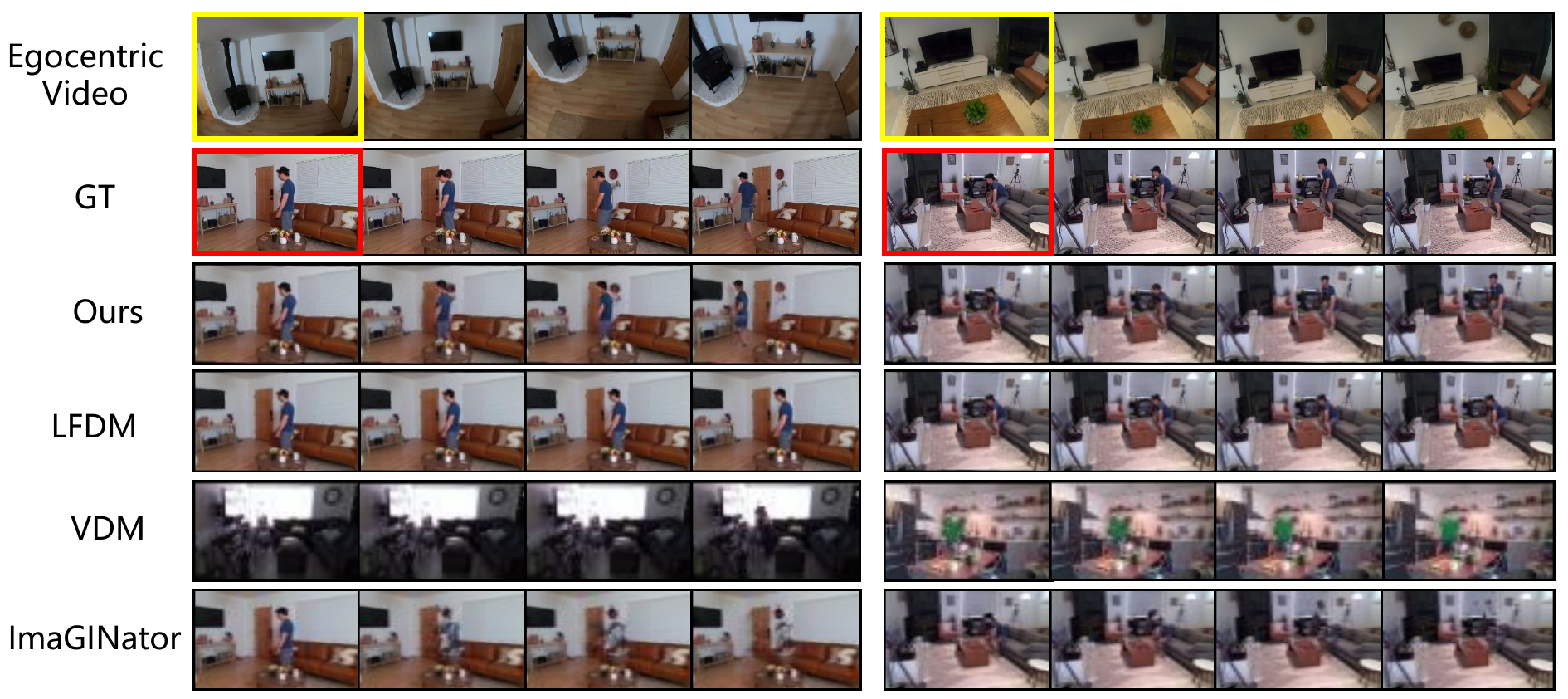}
	\end{overpic}
	\vspace{-10pt}
	\caption{\textbf{The results for exocentric video generation with different methods in \texttt{Unseen} setting.} The yellow box represents the first frame of the egocentric video and the red box represents the first frame of the real exocentric video.}
	\label{result_v2}
\end{figure*}

\begin{figure}[t]
	\centering
		\begin{overpic}[width=0.99\linewidth]{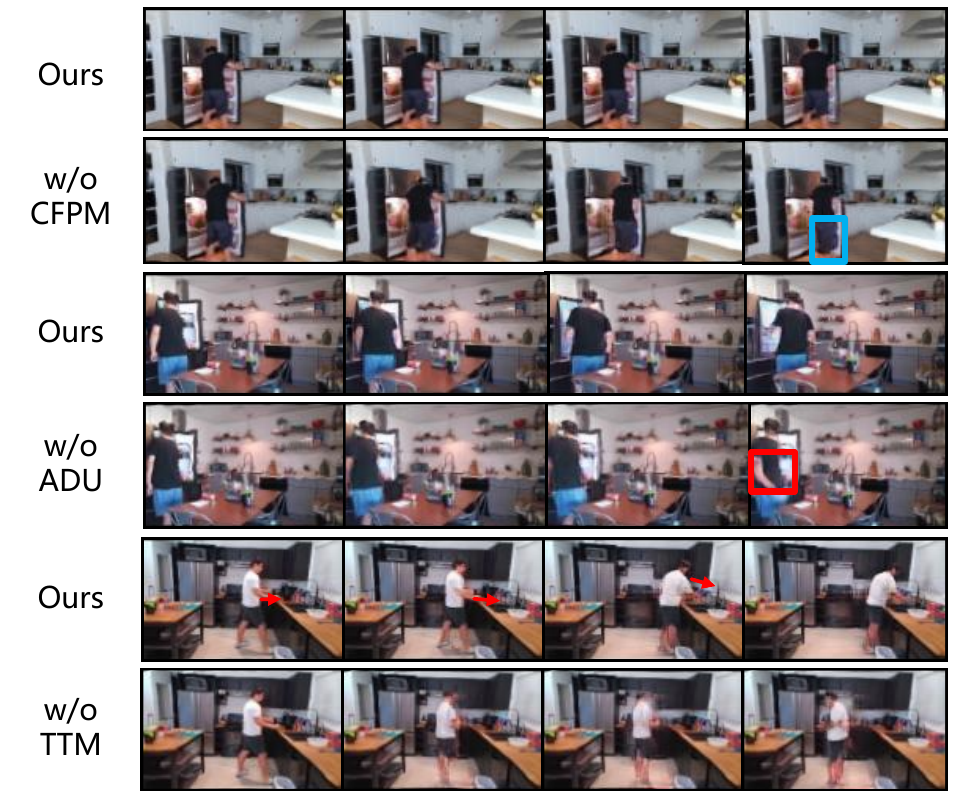}
	\end{overpic}
	\vspace{-8pt}
	\caption{\textbf{Visualization results for ablation study.} We visualize the results of ablation study for the CFPM, ADU and TTM.}
	\label{viz_ab}
 \vskip -0.2in
\end{figure}

\begin{figure}[t]
	\centering
		\begin{overpic}[width=0.99\linewidth]{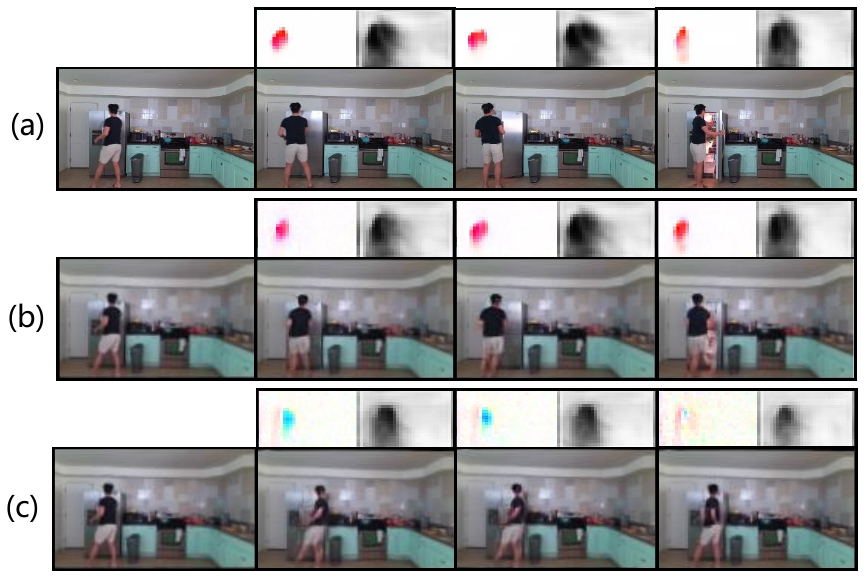}
	\end{overpic}
	\vspace{-8pt}
	\caption{\textbf{Optical flow and occlusion map visualization results.} (a) GT. (b) Ours. (c) w/o TTM.  }
	\label{flow result}
  \vskip -0.2in
\end{figure}

\textbf{Implementation details.\ } In the first stage, the latent optical flow autoencoder follows the structure of LFAE \citep{ni2023conditional} and the optical flow estimation module is implemented with MRAA \citep{siarohin2021motion} and estimates the latent optical flow $f$ and the occlusion map $m$ from the detected object parts. During the training process, we use Adam iterated $120,000$ times at a learning rate of $2e-4$, with the batch size set to $100$ and the size of the input image set to $128 \times 128$. During the second phase, CLIP's visual and text encoders are frozen. The cross-view feature perception module, trajectory transformation module, and the conditional diffusion model are iterated $140,000$ times under training conditions with a learning rate of $2e-4$, batch size set to $10$, input size $128 \times 128$, and the number of video frames $T$ is $24$. %The comparison methods are reproduced according to the open source code, and the number of video frames $T$ is $24$.

\begin{figure}[t]
	\centering
		\begin{overpic}[width=0.98\linewidth]{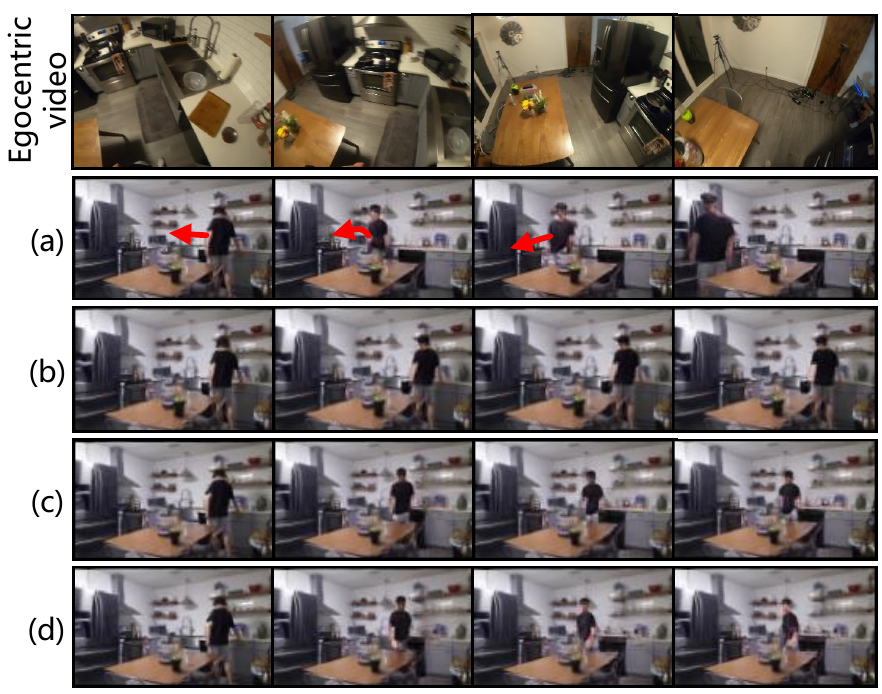}
	\end{overpic}
	\vspace{-7pt}
	\caption{\textbf{The visualization results of different egocentric motion introduction methods.} \textbf{(a)} Ours. \textbf{(b)} Direct input of trajectories as conditions to the diffusion model. \textbf{(c)} Trajectory features spliced with egocentric's class token. \textbf{(d)} Extraction of egocentric using I3D and fusion with exocentric features using cross-attention.}
	\label{result_2}
  \vskip -0.05in
\end{figure}

\begin{table}[t]
\caption{\textbf{Ablation study.} We conduct ablation experiments on the cross-view feature perception module, the action description unit and the trajectory transformation module, respectively.}
   \label{Table:Ablation study}
\vskip 0.15in
\begin{center}
\renewcommand{\arraystretch}{1.}
  \renewcommand{\tabcolsep}{10.5pt}
\begin{small}
%\begin{sc}
 \begin{tabular}{c||ccc}
    \hline
    \Xhline{2.\arrayrulewidth}
       &      $\text{LPIPS} \downarrow$ & $\text{FVD} \downarrow$  & $\text{KVD} \downarrow$ \\ 
    \hline
\Xhline{2.\arrayrulewidth}
    w/o CFPM    & $0.140$ & $309.34$ & $30.42$  \\
    w/o ADU   	  & $0.202$ & $284.46$  & $29.68$  \\
    w/o TTM & $0.223$ &  $332.98$ & $34.56$ \\
    \hline
    \rowcolor{mygray}
    Ours &  $\textbf{0.133}$ & $\textbf{219.81}$ & $\textbf{22.49}$   \\
\hline
    \Xhline{2.\arrayrulewidth}
\end{tabular}
%\end{sc}
\end{small}
\end{center}
\vskip -0.2in
\end{table}

\subsection{Quantitative and Qualitative Comparisons}
The experimental results are shown in Table \ref{Table:total result}, our method outperforms all the metrics on resolutions of $128$ and $64$. In the LIPIS metric, our method exceeds the other methods, indicating that our model generates higher quality images closer to the real exocentric videos. On the other hand, on the two metrics of FVD and KVD for evaluating the video, the advantage of our method over other models is more significant. It suggests that our model, guided by head trajectories, makes the generated video segments exhibit more reasonable spatial and temporal coherence in human motion. Meanwhile, we also show the generation results of the different methods (as shown in Fig. \ref{result} and Fig. \ref{result_v2}). From the first example in Fig. \ref{result}, our approach accurately generates exocentric videos that align spatially with the movements depicted in the egocentric video. This underscores the network's ability to establish a meaningful connection between exocentric and egocentric frames, enabling precise perception of movement states in the egocentric perspective and generating relevant exocentric video sequences. In Fig. \ref{result_v2}, the first example illustrates that our method consistently produces more realistic motion sequences and body poses closely resembling the real video, even with increased spatial location variations. It underscores the effectiveness of our cross-view feature perception module, facilitating a connection between ego-to-exo views and transferring it to exocentric features through the trajectory transformation module. This guidance ensures that the network generates accurate motion directions, and the inclusion of action description enhances pose accuracy.

\subsection{Ablation Study}
Table \ref{Table:Ablation study} shows the ablation study results for the CFPM, the TTM, and the ADU. It reveals that egocentric head motion trajectory significantly influences model performance, providing crucial cues for human movement's direction and relative translational speeds. In contrast, the impact of action description is relatively small, primarily enhancing accuracy in interaction cues with minimal influence on the overall results. Fig. \ref{viz_ab} shows the results of video generation for the ablation study. Without the CFPM, the model relies on the rough head motion information to generate the corresponding video. It does not consider the relative spatial relationship between the human and the object in the exocentric image, leading to the generation of a video that does not correspond to the actual human motion. When there is no input from the action description condition, it is possible to generate anomalous human interaction. When there is no head trajectory, the generated video human movement may be more confusing, have no clear direction, and not be coherent in temporal. Further, we visualize the optical flow and occlusion map generated by our method and the model without the head trajectory included, as shown in Fig. \ref{flow result}. Our model can migrate motion cues to exocentric features based on the cues provided by the head motion trajectory and the connection between egocentric and exocentric, resulting in more accurate generated optical flow and occlusion maps.

\subsection{Performance Analysis}
\textbf{Different motion information fusion ways.} Table \ref{Table:Ablation studyv2} shows the impact of different approaches incorporating egocentric motion information on model performance. ``Conditional input $\mathcal{T}$'' involves directly introducing the trajectory into the diffusion model as a condition, while ``$\mathcal{T}$ + ego cls concat'' denotes concatenating trajectory information with the egocentric class token,  and then is then fused through the transformer layer. ``Egocentric video'' indicates that egocentric features are extracted using I3D \citep{carreira2017quo} and combined with the egocentric features. Meanwhile, we visualize the results of the exocentric video generation, as shown in Fig. \ref{result_2}. Though the egocentric videos with large scene changes, \ie the direction and relative position of the human motion changes, our method can still provide a superior reproduction of the human motion in the exocentric viewpoint. With ``Conditional input $\mathcal{T}$'', directly inputting the head motion trajectory into the diffusion model as a condition, the model struggles to align the motion in exocentric and egocentric views, resulting in challenges in generating accurate video sequences. For the case of ``$\mathcal{T}$ + ego cls concat'', due to the large gap between the class  token and the trajectory features, directly concatenating the trajectory with the egocentric features using the transformer layer does not guarantee that it is effective in modeling the temporal dynamics of the egocentric features, which leads to generating egocentric videos with a large range of rotations and movements. For the ``egocentric video'', redundant information in the egocentric video results in extracted features containing more spatial content. It introduces irrelevant information when transferring egocentric motion to exocentric features, hindering the generation of videos depicting extensive movement and larger-angle rotations.

\begin{table}[t]
\caption{\textbf{Different methods of introducing egocentric motion.} We explored different ways of introducing egocentric motion information into the model. }
   \label{Table:Ablation studyv2}
%\vskip 0.15in
\begin{center}
\renewcommand{\arraystretch}{1.}
  \renewcommand{\tabcolsep}{8pt}
\begin{small}
%\begin{sc}
 \begin{tabular}{c||ccc}
    \hline
    \Xhline{2.\arrayrulewidth}
       &      $\text{LPIPS} \downarrow$ & $\text{FVD} \downarrow$  & $\text{KVD} \downarrow$ \\ 
    \hline
\Xhline{2.\arrayrulewidth}
    conditional input  $\mathcal{T}$ & $0.187$ & $283.40$ & $28.96$ \\
 $\mathcal{T}$ + ego cls concat & $0.181$  & $295.17$ & $37.37$ \\
 egocentric video & $0.174$ & $275.82$ & $32.80$ \\
 \hline
    \rowcolor{mygray}
   Ours &  $\textbf{0.133}$ & $\textbf{219.81}$ & $\textbf{22.49}$    \\
\hline
    \Xhline{2.\arrayrulewidth}
\end{tabular}
%\end{sc}
\end{small}
\end{center}
\vskip -0.1in
\end{table}

\begin{figure}[t]
	\centering
		\begin{overpic}[width=0.9\linewidth]{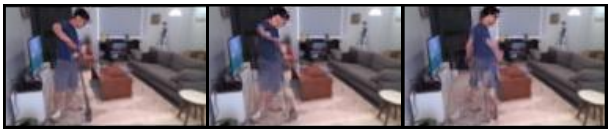}
	\end{overpic}
	\vspace{-10pt}
	\caption{\textbf{Failure Case.} It is challenging to learn the connection between movement patterns between ego-to-exo when the egocentric movement is slight in amplitude.}
	\label{fc}
 \vskip -0.2in
\end{figure}

\textbf{Different frames.}  We examine the model performance 
\begin{wraptable}{r}{0.276\textwidth}
\vskip -0.2in
\caption{\textbf{Different frames.} We test the results of IDE for predicting videos with different frame numbers.}
\vspace{5pt}
  \centering
  \renewcommand{\arraystretch}{1.}
  \renewcommand{\tabcolsep}{5pt}
   %\footnotesize
   \scriptsize
   %\vspace{-2mm}
   \label{Table:different frames}
  \begin{tabular}{c|ccc}
   \hline
    \Xhline{2.\arrayrulewidth}
  Num &  $\text{LPIPS} \downarrow$ & $\text{FVD} \downarrow$  & $\text{KVD} \downarrow$ \\ \hline
  $16$ & $0.205$  & $167.29$ & $15.85$ \\
  $24$ & $0.133$ & $219.81$ & $22.49$ \\
  $32$ &  $0.115$ & $304.91$ & $65.43$ \\
  $40$ &  $0.120$ & $423.95$ & $142.37$ \\
  \hline
    \Xhline{2.\arrayrulewidth}
    \end{tabular} 
    \vspace{-4.mm}
    \end{wraptable}
for generating different video lengths, as shown in Table \ref{Table:different frames}. As the number of frames increases, the model slightly decreases in the FVD and KVD metrics for evaluating video quality but still maintains well-predicted performance, with the model at $32$ frames being close in FVD to the other models at $24$ frames, which suggests that our model maintains temporal and spatial consistency in predicting longer video sequences.

\textbf{Limitations.} Fig. \ref{fc} illustrates a failure case. With minimal head motion, a significant inconsistency arises between head and human movement. Establishing a connection between exocentric and egocentric motion becomes challenging, resulting in a blurred or abnormal generation effect. Future improvements may involve a more in-depth analysis of the scene content and leveraging accurate information about the scene to enhance exocentric video generation.

\section{Conclusion}
In this paper, we propose an \textbf{I}ntent-\textbf{D}riven \textbf{E}go-to-exo video generation framework (\textbf{IDE}) that leverages human movement along with action descriptions to jointly represent the human action intention and is employed to facilitate the exocentric video generation. We conduct experiments on the relevant dataset containing rich ego-exo video pairs, and IDE outperforms state-of-the-art video generation models, demonstrating the effectiveness of our method.

\textbf{Broader lmpacts. } The research on ego-exo video generation will advance the realization of AR/VR. However, exocentric videos of bloody and violent actions will have a stronger visual impact and should be prevented and filtered.

\bibliography{example_paper}
\bibliographystyle{icml2024}

%%%%%%%%%%%%%%%%%%%%%%%%%%%%%%%%%%%%%%%%%%%%%%%%%%%%%%%%%%%%%%%%%%%%%%%%%%%%%%%
%%%%%%%%%%%%%%%%%%%%%%%%%%%%%%%%%%%%%%%%%%%%%%%%%%%%%%%%%%%%%%%%%%%%%%%%%%%%%%%
% APPENDIX
%%%%%%%%%%%%%%%%%%%%%%%%%%%%%%%%%%%%%%%%%%%%%%%%%%%%%%%%%%%%%%%%%%%%%%%%%%%%%%%
%%%%%%%%%%%%%%%%%%%%%%%%%%%%%%%%%%%%%%%%%%%%%%%%%%%%%%%%%%%%%%%%%%%%%%%%%%%%%%%

%%%%%%%%%%%%%%%%%%%%%%%%%%%%%%%%%%%%%%%%%%%%%%%%%%%%%%%%%%%%%%%%%%%%%%%%%%%%%%%
%%%%%%%%%%%%%%%%%%%%%%%%%%%%%%%%%%%%%%%%%%%%%%%%%%%%%%%%%%%%%%%%%%%%%%%%%%%%%%%

\end{document}